\documentclass[letterpaper, 10 pt, conference]{ieeeconf}  %

\IEEEoverridecommandlockouts                              %

\usepackage[english]{babel}

\usepackage{graphicx}
\usepackage{animate}
\usepackage{epsfig} 				%
\usepackage{epstopdf}

\usepackage{listings}
\usepackage{color}
\usepackage{nameref}
\usepackage{hyperref}
\usepackage[colorinlistoftodos]{todonotes}
\usepackage{amsmath}	 			%
\usepackage{amssymb}  				%

\usepackage{dsfont}			%
\usepackage{mathtools}

\usepackage{epigraph}
\usepackage{lscape}
\usepackage[]{nomencl}				%
\usepackage{algorithm}
\usepackage{algorithmic}
\usepackage{multicol}
\usepackage{multirow}
\usepackage{etoolbox}

\usepackage{caption}
\usepackage{subcaption}

\usepackage{wrapfig}
\usepackage{multimedia}

\usepackage{siunitx}
\usepackage{flushend}

\usepackage{floatflt}

\usepackage{url}
\usepackage[absolute,overlay]{textpos}

\usepackage{bibentry}

\usepackage{tikz}
\usepgflibrary{arrows}	%

\usepackage{float}
\usepackage[utf8]{inputenc}
\usepackage[english]{babel}
\usepackage{graphics}
\usepackage{animate}

\usepackage{listings}

\usepackage[nostamp]{draftwatermark}

\usepackage{extarrows}		%

\allowdisplaybreaks[1]

\newcommand{\playvideo}[1]{\href{run:#1}{\includegraphics[scale=0.12]{\RCPath fig/empty}}}

\usepackage{lipsum}
\usepackage{framed}

\newcommand{\email}[1]{\href{mailto:#1}{\nolinkurl{#1}}}

\newcommand{\link}[1]{\colora{\url{#1}}}

\newcommand{\fig}[1]{Figure~\ref{#1}}

\newcommand{\app}[1]{Appendix~\ref{#1}}
\newcommand{\tab}[1]{Table~\ref{#1}}

\definecolor{matlab1}{rgb}{0,0,1}
\definecolor{matlab2}{rgb}{0,0.5,0}
\definecolor{matlab3}{rgb}{1,0,0}
\definecolor{matlab4}{rgb}{0,0.75,0.75}
\definecolor{matlab5}{rgb}{0.75,0,0.75}
\definecolor{matlab6}{rgb}{0.75,0.75,0}
\definecolor{matlab7}{rgb}{0.25,0.25,0.25}

\definecolor{darkgreen}{rgb}{0,0.5,0}		%
\definecolor{purple}{rgb}{0.75,0,0.75}
\definecolor{pink}{rgb}{1,0.4,0.6}

\newcommand{\capitalize}[1]{\expandafter\MakeUppercase\expandafter{#1}}

\newcommand{\colora}[1]{{\usebeamercolor[fg]{framesubtitle}#1}}

\makeatletter
\newcommand*{\compress}{\@minipagetrue}
\makeatother

\renewcommand{\vec}[1]{\boldsymbol{#1}}				%

\newcommand{\q}{\vec{q}}					%
\ifdef{\dq}{\renewcommand{\dq}{\dot{\q}}}{\newcommand{\dq}{\dot{\q}}}

\usepackage{array}
\makeatletter
\newcommand{\thickhline}{%
    \noalign {\ifnum 0=`}\fi \hrule height 1pt
    \futurelet \reserved@a \@xhline
}
\newcolumntype{"}{@{\hskip\tabcolsep\vrule width 1pt\hskip\tabcolsep}}
\makeatother
 
\AtEndDocument{\par\leavevmode}

\newcommand{\citep}[1]{\cite{#1}}

\title{\LARGE \bf
Manipulation by Feel:\\ Touch-Based Control with Deep Predictive Models
}

\author{Stephen Tian$^{\star,1}$, Frederik Ebert$^{\star,1}$, Dinesh Jayaraman$^1$, Mayur Mudigonda$^1$,\\ Chelsea Finn$^1$, Roberto Calandra$^2$, Sergey Levine$^1$%
\thanks{$^\star$ Equal contribution}%
\thanks{This work was  supported by Berkeley DeepDrive, NSF IIS-1614653, and the Office of Naval Research (ONR)}%
\thanks{$^{1}$Department of Electrical Engineering and Computer Sciences, University of California, Berkeley, USA\newline
        {\tt\small \{stephentian, febert, dineshjayaraman, mudigonda, cbfinn, sergey.levine\}@berkeley.edu}}%
\thanks{$^{2}$Facebook AI Research, Menlo Park, CA, USA\newline
        {\tt\small rcalandra@fb.com}}%
}

\begin{document}

\maketitle
\begin{abstract}
Touch sensing is widely acknowledged to be important for dexterous robotic manipulation, but exploiting tactile sensing for continuous, non-prehensile manipulation is challenging. General purpose control techniques that are able to effectively leverage tactile sensing as well as accurate physics models of contacts and forces remain largely elusive, and it is unclear how to even specify a desired behavior in terms of tactile percepts. In this paper, we take a step towards addressing these issues by combining high-resolution tactile sensing with data-driven modeling using deep neural network dynamics models.
We propose deep tactile MPC, a framework for learning to perform tactile servoing from raw tactile sensor inputs, without manual supervision. We show that this method enables a robot equipped with a GelSight-style tactile sensor to manipulate a ball, analog stick, and 20-sided die, learning from unsupervised autonomous interaction and then using the learned tactile predictive model to reposition each object to user-specified configurations, indicated by a goal tactile reading.
Videos, visualizations and the code are available here:

\footnotesize{\url{https://sites.google.com/view/deeptactilempc}}
 \end{abstract}

\section{Introduction}

Imagine picking up a match stick and striking it against a matchbox to light it, a task you have performed with ease many times in your life. 
But this time, your hand is numb. 
In 2009, Johansson et al.~\citep{Johansson2009} performed this experiment, studying the impact of anesthetizing the fingertips of human subjects on their ability to perform this task. 
The results were striking:
human subjects could barely manage to pick up a match stick, let alone light it. 
Videos of this experiment show human clumsiness~\citep{matchstick_experiment} that is strikingly reminiscent of the faltering, lurching struggles of modern robots~\citep{robots_falling}.

Why did taking away the sensation of touch have such an impact on these subjects? 
Touch is unique among sensory modalities in that it is physically immediate and permits direct measurement of ongoing contact forces during object interactions, from which it is possible to infer friction, compliance, mass, and other physical properties of surfaces and objects. 
This knowledge is critical for manipulation tasks like matchstick striking. 
Visual sensing is a poor substitute: not only is it physically remote, but it is also usually occluded by the actuators at the points of contact.
Manipulation without touch is perhaps akin to navigation without vision. 

\begin{figure}[t]
  \centering
  	\begin{subfigure}[b]{0.49\linewidth}
		\centering
		\includegraphics[width=\linewidth]{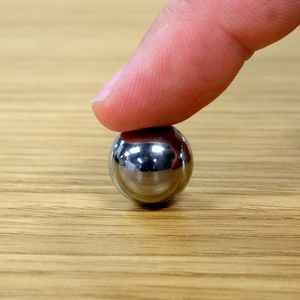}
	\end{subfigure}
	\hfill
	\begin{subfigure}[b]{0.49\linewidth}
		\centering
		\includegraphics[width=\linewidth]{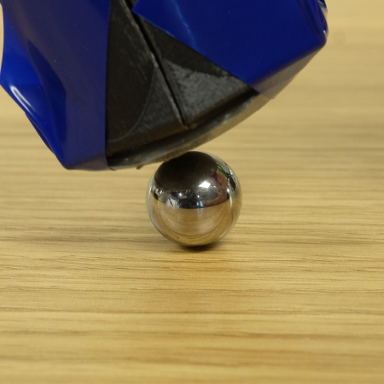}
	\end{subfigure}
  \caption{ (\textit{Left})~For fine manipulation, humans rely mostly on touch, as vision is occluded by the finger itself. (\textit{Right})~Our custom-built GelSight touch sensor. We train a video prediction model on the tactile modality, and use this model to perform object repositioning tasks.}
  \label{fig:teaser}
\end{figure}
The importance of tactile sensing has long been acknowledged in the robotics community~\cite{yousef2011tactile,dahiya2010tactile,hammock201325th}, but exploiting touch in robots has proven exceedingly challenging for three key reasons: (i) tactile sensing technology has largely been limited to sparse pointwise force measurements, a far cry from the rich tactile feedback of biological skin, (ii) accurate physics models of contacts and forces have remained elusive, and (iii) it is unclear how to even specify a desired tactile goal, such as holding a matchstick, or striking it obliquely against the matchbox. 

In this paper, we show the promise of robotic control with rich tactile feedback, by exploiting recent advances to tackle
these  difficulties. 
Firstly, deformable elastomer-based tactile sensing (e.g., GelSight) has proven to be an extremely versatile, high-bandwidth, high-spatial resolution alternative to traditional tactile sensing~\cite{Yuan2017b,Calandra2017}. 
This provides our first stepping stone.
Secondly, while high-resolution touch sensing such as GelSight (see \fig{fig:teaser} and \ref{fig:tasks}) provides us with adequate sensory signals, we also need a control mechanism that can handle such high-dimensional observations and choose intelligent actions~\cite{Mnih2015,Levine2016}. Our second stepping stone comes from the deep learning literature, where deep models of high-dimensional data have been successfully employed to enable a variety of vision-based robotic skills~\cite{foresight,savp,Ebert2017}.

Finally, as we will show, the combination of these two advances makes it feasible to plan towards tactile goals specified directly in the raw tactile observation space. 
Goal specification in this manner is not only much more informative than, say, in the space of forces at sparse contact points, but is also often much more natural for a user to specify.

Concretely, our contributions are as follows. We train deep model-based control policies that operate directly on observed raw high-dimensional tactile sensing maps.
We show that such policies may be learned entirely without rewards, through diverse unsupervised exploratory interactions with the environment. Finally, we show how the desired manipulation outcomes for our policies may be specified as goals directly in the tactile observation space.
We demonstrate and evaluate these contributions on a high-precision tactile ball rolling task, a joy-stick re-positioning task and a die rolling task: a robot arm with three linear axes is equipped with a tactile sensor at its end effector. 
Its goal is to move the end-effector into a configuration so that a desired tactile goal-pattern is measured with the sensor.

These tasks are designed to exhibit two key difficulties that are shared by a wide range of manipulation tasks: (i) An external object must be in contact with the robot end-effector in a specific desired configuration. This is a common feature of many manipulation and tool use problems. 
For example, when trying to light a match, a specific sequence of contact states needs to be attained. (ii) While performing the task, the object of interest, such as the ball bearing, becomes occluded from view, therefore controllers or policies that only have access to a remote visual observation are unable to solve the task.
In the experiments for these three distinct manipulation tasks, our method outperforms hand-designed baselines for each task. 
We see these results as an important step towards integrating touch into solutions for general robotic manipulation tasks.

\section{Related Work}
\label{sec:related}

Prior work on touch-based control has proposed methods ranging from manual design of control laws~\cite{sikka1994tactile} to extracting and controlling high-level features from touch sensors~\citep{Li2013,Lepora2017}. In contrast to these methods, our approach does not rely on pre-specified control laws and features. We learn a general-purpose predictive model that can be used to accomplish a variety of tasks at test time. Furthermore, we use a high-resolution touch sensor based on the GelSight design~\cite{johnson2011microgeometry}, which provides detailed observations of the contact surface in the form of a camera image.

Prior work has also used reinforcement learning to learn to stabilize an object with touch sensing~\cite{vanhoofstable} and explored learning forward predictive models for touch sensors~\citep{Veiga2017}.
While this prior work used low-dimensional readings from a BioTac sensor, our model operates directly on the raw observations of a GelSight sensor, which in a our case is an RBG image downsampled to 48x64 pixels. 
We demonstrate that comparatively high-resolution tactile sensing in conjunction with the proposed tactile MPC algorithm allows to reposition freely-moving objects according to user-specified goals, a more complex task than those demonstrated in prior work~\cite{vanhoofstable,Veiga2017}. To address this high-dimensional prediction problem, we build on recent work on control via video prediction~\cite{foresight,Ebert2017}. 
Prior work in this area has used video prediction in combination with model-predictive control to perform non-prehensile object repositioning from RGB camera images. To our knowledge, no prior work has used video prediction models together with touch sensing for touch-based object repositioning.
Concurrent work \cite{sutanto2018learning} learned a two dimensional latent space and dynamics model to perform control for following human demonstrations given in the tactile space, however handling of objects has not been shown yet.

A variety of different touch sensor designs have been proposed in the literature \citep{yousef2011tactile}, though affordability, sensitivity, and resolution all remain major challenges. The BioTac~\cite{fishel2013syntouch,fishel2012sensing} sensor has been widely used in robotics research, particularly for grasping~\citep{Chebotar2016,matulevich2013utility}, but it provides only a limited number of measuring channels (i.e., 19 or 22 for different configurations).
The GelSight design, which consists of a camera that observes deformations in a gel, offers good resolution, though at the cost of latency~\cite{johnson2011microgeometry}. In our case, this tradeoff is worthwhile, because the resolution
of the sensor allows us to precisely reposition objects in the finger.
GelSight-style sensors have been used in a number of prior works for other applications, including tracking~\cite{izatt2017tracking}, inserting USB connectors~\cite{li2014localization}, estimating object hardness~\cite{yuan2017shape}, and grasping~\cite{Calandra2018}. 
To our knowledge, our work is the first to employ them for object repositioning with learned predictive models.

\section{Tasks and Hardware Setup}\label{sec:hardware}

\begin{figure}[t]
  \centering
\includegraphics[width=0.31\linewidth]{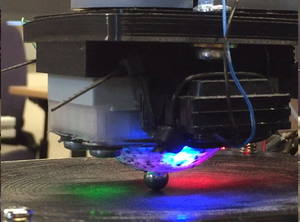}  
\includegraphics[width=0.31\linewidth]{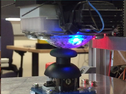}
\includegraphics[width=0.31\linewidth]{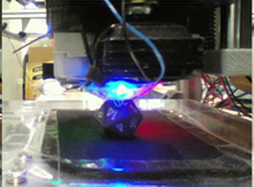}  
  \caption{We evaluate deep tactile MPC on 3 different fine-grained manipulation tasks: (left to right) ball repositioning, joystick deflection, and die rolling to reach a specified face.}
  \label{fig:tasks}
\end{figure}

While our aim is to develop control approaches for general manipulation, we focus on three representative tactile control tasks: rolling a ball to a specified position, manipulating an analog stick from a game controller, and rolling a die to a specified face, see \fig{fig:tasks}. Each of these tasks presents unique challenges, making them well-suited for evaluating our approach. In this section, we describe the tasks, the sensor, and the experimental hardware setup.

\begin{figure*}[t]
  \centering
  \includegraphics[width=\linewidth]{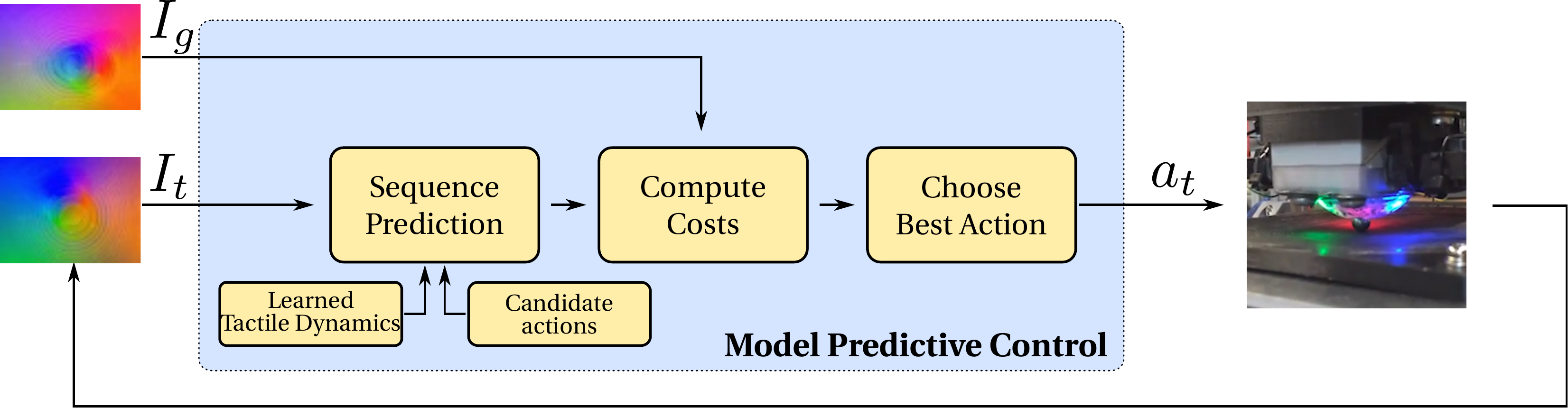}
  \caption{Deep tactile model predictive control: given the current tactile observation and a learned deep predictive model, we can predict the outcomes for different action sequences. We use this model within a model predictive control algorithm based on stochastic optimization. At each time-step the algorithm samples multiple potential action sequences and computes their cost, which depends on the difference between the predicted tactile observations and the goal observation. The first action of the actions sequence that attained lowest cost is then applied to the actuators.}
  \label{fig:control}
\end{figure*}

\paragraph{Ball repositioning task} The ball repositioning task requires the robot to move a ball bearing to a target location on the sensor, which requires careful modulation of the contact force between the ball and the finger without the ball slipping out of the finger.
Since the ball is underactuated, properly balancing the forces is critical.

\paragraph{Analog stick deflection task} The second task requires the robot to deflect an analog stick. This task presents an additional challenge: the robot must intentionally break and reestablish contact to deflect the analog stick in the desired direction, since sliding the finger along the stick will deflect it in undesirable ways. We encourage the reader to view the supplementary video for an illustration. The only way to perform the task successfully is to lift the finger off of the stick by moving vertically, repositioning it, bringing it back down onto the stick, and then deflecting the stick in the desired direction. This adds the difficulty of making the system only partially observable (when the finger loses contact with the joystick), and requires the model to acquire a more fine grained understanding of contact dynamics.

\paragraph{Rolling a 20-sided die} This third task requires the robot to roll a 20-sided die so that a specified die face is facing upwards.
This task was chosen for two reasons: First this task has a comparably high-level of difficulty due to slippage and undesired rolling of the die. Second this task allows us to use a simple and intuitive success metric --- the fraction of trails where the desired face ended up on top.

\paragraph{Tactile sensor} For tactile sensing, we use a custom elastomer-based sensor based on the GelSight design~\cite{Yuan2017b} with a diameter of \SI{4}{\centi\meter}. 
A standard webcam is embedded directly into the Gel producing high-resolution images of the surface deformations of the elastomer. Example images are shown in \fig{fig:Video-predictions}. Our choice of tactile sensor is key to our approach for the two reasons: (i) it allows us to use comparatively high-resolution observations, which aids both in control, as well as in setting expressive self-supervised goals at test time (as we will show), and (ii) it naturally offers a compliant surface at the point of contact, which is important in many manipulation settings including ours.

\begin{figure}[h]
  \centering
  \includegraphics[width=0.98\linewidth]{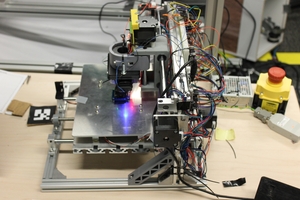}
  \caption{Hardware setup. Custom manufactured GelSight sensor mounted on a modified 3-axis CNC machine, which allows for linear translation along each of the three axes.}
  \label{fig:hardware}
 \end{figure}

\paragraph{Hardware setup} In order to study tactile control with our sensor, we mount it on a modified miniature 3-axis CNC machine (see \fig{fig:hardware} in the appedix). This machine has a precision of \SI{\approx 0.04}{\milli\meter}, which allows it to reposition the sensor accurately based on the actions commanded by our controller.

\paragraph{Autonomous data collection} To train our deep predictive model, we need to collect training data of the robot interacting with its environment. We autonomously collected 7400 trajectories for the ball, around 3000 trajectories for the analog stick, and 4500 trajectories for the die experiment. Each trajectory consists of 15 to 18 time steps, depending on the experiment. These training trajectories were collected by applying random movements along each of the three axes. More details about the data collection process are provided in \app{sec:app_data_coll}.

\section{Deep Tactile Model-Predictive Control}

The use of a high-resolution sensor such as the GelSight enables fine control, but also presents a major challenge: the high-dimensional observation space makes modeling and control substantially more difficult. To allow performing a variety of different manipulation tasks at test-time using a large dataset collected beforehand, we explore a model-based method for touch-based control. Our method builds on prior work on control via visual prediction~\cite{foresight,Ebert2017}. In this class of methods, a deep recurrent convolutional network is trained to predict future video frames conditioned on the most recent observations and a sequence of future actions. More information about the model are provided in \app{sec:app_model}. At test time, this model can be used to perform a variety of manipulation tasks by optimizing over the actions until the model produces the predictions that agree with the user's goal, and then executing those actions on the robot. Prior work has applied this approach to vision-based object repositioning~\cite{foresight,Ebert2017}.

\begin{figure}[t]
  \centering
  \includegraphics[width=0.98\linewidth]{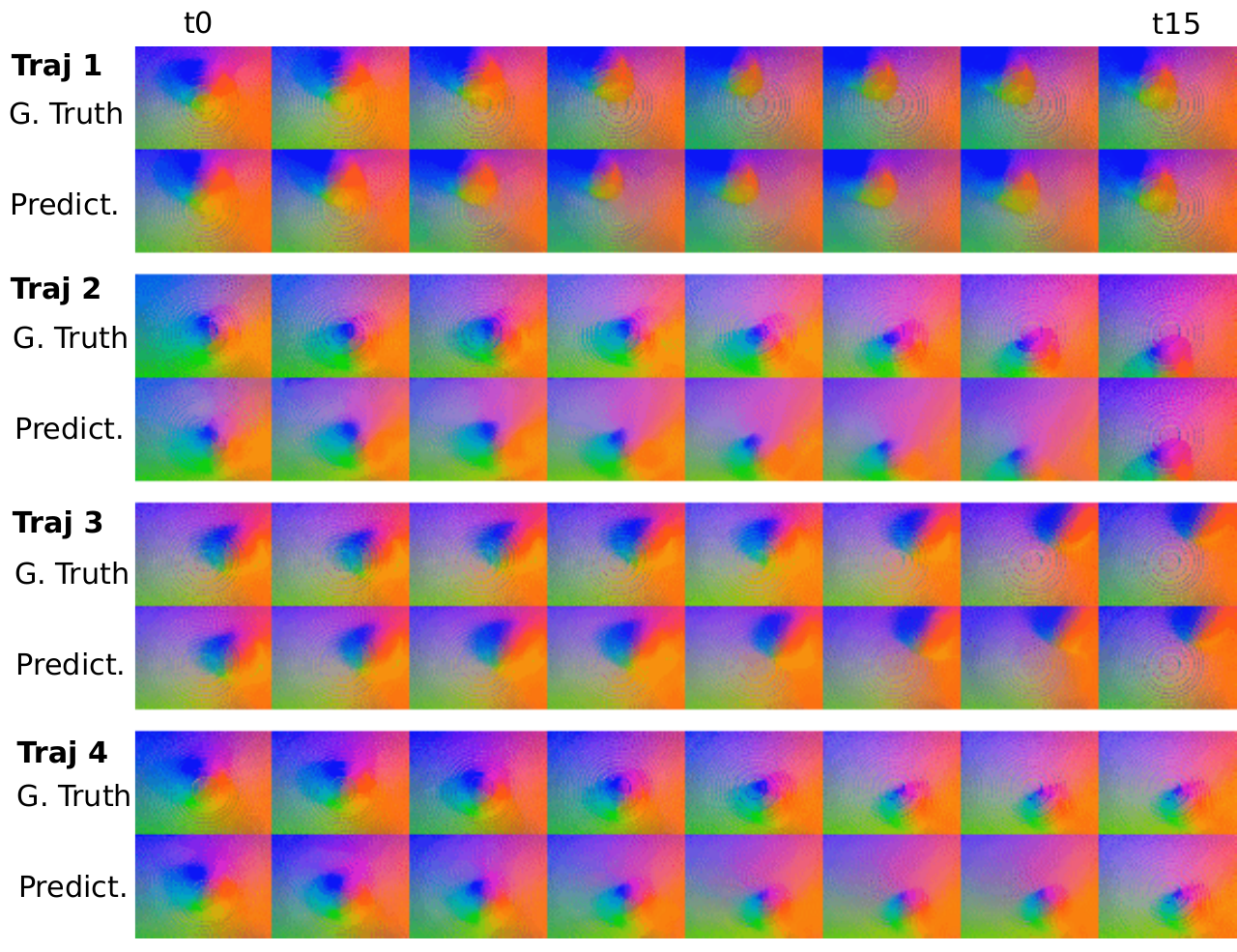}
  \caption{ \small{Four different predicted sequences for the ball bearing task, conditioned on images and actions from the test set: the top film strip in each row shows the ground truth observations, the bottom one shows the predictions made by our model when conditioned on the same action sequence. The actions consist of movements between 0 and \SI{2.8}{\milli\meter} in length along the horizontal axes, and between 0 and \SI{0.4}{\milli\meter} in length along the vertical axis.}}
  \label{fig:Video-predictions}
\end{figure}

\paragraph{Deep predictive model} The particular model that we use has the same architecture as the video-prediction models proposed in prior work~\cite{foresight,Ebert2017}. Concretely, we use the architecture proposed in~\cite{savp}, and train it to predict future GelSight sensor observations $\hat{I}_{1:T} \in \mathbb{R}^{T \times H\times W \times 3}$  conditioned on the current observation $I_0$ and a sequence of candidate actions $a_{1:T}$, where $T$ denotes the prediction horizon.\footnote{While the architecture was proposed to generate stochastic predictions in~\cite{savp}, we train a deterministic variant instead.} This predictive model can be written as $\hat{I}_{1:T} = g(a_{1:T}, I_0)$. In Figure \ref{fig:Video-predictions} we show several example predictions of our model on test-set trajectories. We can see that the model accurately predicts the contact pattern for a sequence of 13 time-steps into the future.

\paragraph{Goal specification} At test-time, the user specifies a goal by providing a goal \emph{tactile image}: a reading from the GelSight sensor for the desired configuration, which we denote as $I_g$. While a number of methods could be used to specify goals, this approach is simple and general, and allows us to evaluate our method on a ``tactile servoing'' task. 

\begin{figure}[t]
  \centering
  \includegraphics[width=\linewidth]{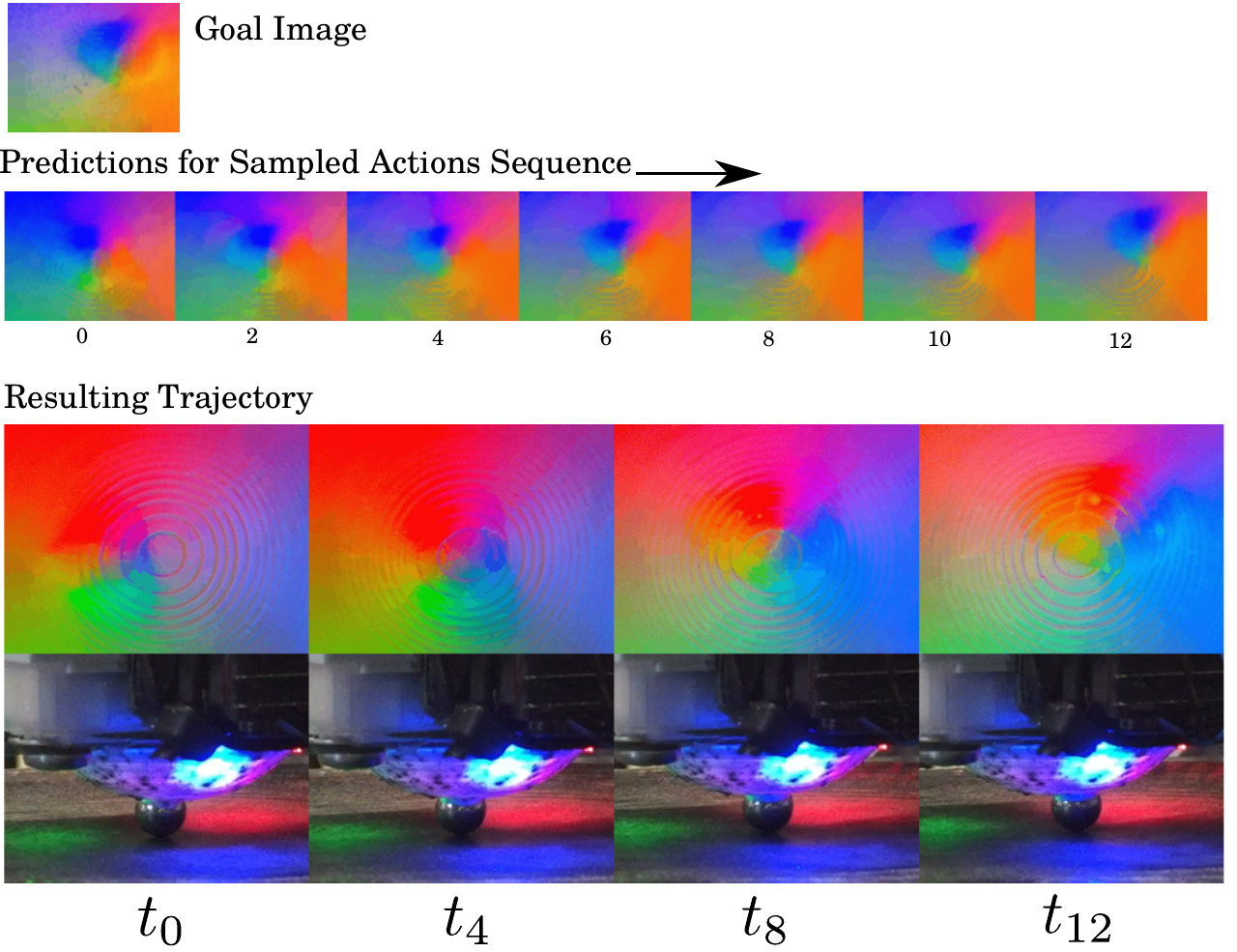}
  \caption{\small Example rollout for the ball-bearing task. The goal is to reach the the goal-image in the top row. The predicted frames for the action sequence that achieved lowest cost is shown in the second row, only every second prediction step is shown. The third row shows the actual trajectory taken by the robot for both the tactile image and side image.}
  \label{fig:planning}
\end{figure}

\paragraph{Tactile MPC control} Once the predictive model has been trained, we may use it to plan to achieve any user-specified goal configuration  $I_g$ in the tactile observation space. For this, we employ model-predictive control with the learned predictive model. We use an optimization-based planner to optimize over the action sequence at each time step to determine the actions for which the predicted outcome is closest to goal tactile image $I_g$, as illustrated in \fig{fig:control}.
The planning problem is formulated as the minimization of a cost function $c_t(I_g, \hat{I}_t)$ which provides a distance metric between the predicted image $\hat{I}_t$ and the goal image $I_g$. In this work we set $c(\cdot, \cdot)$ to the mean squared error (MSE) in pixel-space between $I_g$ and $\hat{I}_t$, such that the optimization is given by
\begin{equation}
a_{1:T}  = \arg\min_{a_{1:T}} \sum_{t = 1, \dots, T} c_t(I_g, \hat{I}_t)\,,
\label{eq:cost}
\end{equation}
where $c_t \in \mathbb{R}$. We perform sampling-based planning using the cross-entropy method (CEM) \cite{cem-rk-13}. To compensate for inaccuracies in the model, the action sequences are recomputed at each time step $t \in \{0,...,t_{max}\}$ following the framework of model-predictive control (MPC). At each real-world step $t$, the first action of the best action sequence is executed on the robot. The planning process is illustrated in \autoref{fig:control}. \autoref{fig:planning} shows the executing of tactile MPC on the ball-bearing task.

\paragraph{Implementation details} We use three CEM iterations for optimization, with 100 samples each. The prediction horizon for the video-prediction model is between 15 and 18 depending on the task. Each action is repeated three times, such that the plan consists of five or six actions. This planning horizon is usually sufficient to reach the goal from most configurations considered in our experiments. Using time-correlated actions helps to reduce the search space for the optimizer, thus reducing the number of required samples and increasing the control rate.

\section{Experimental Results}
\label{sec:results}

\begin{figure}[t]
  \centering
  \includegraphics[width=\linewidth]{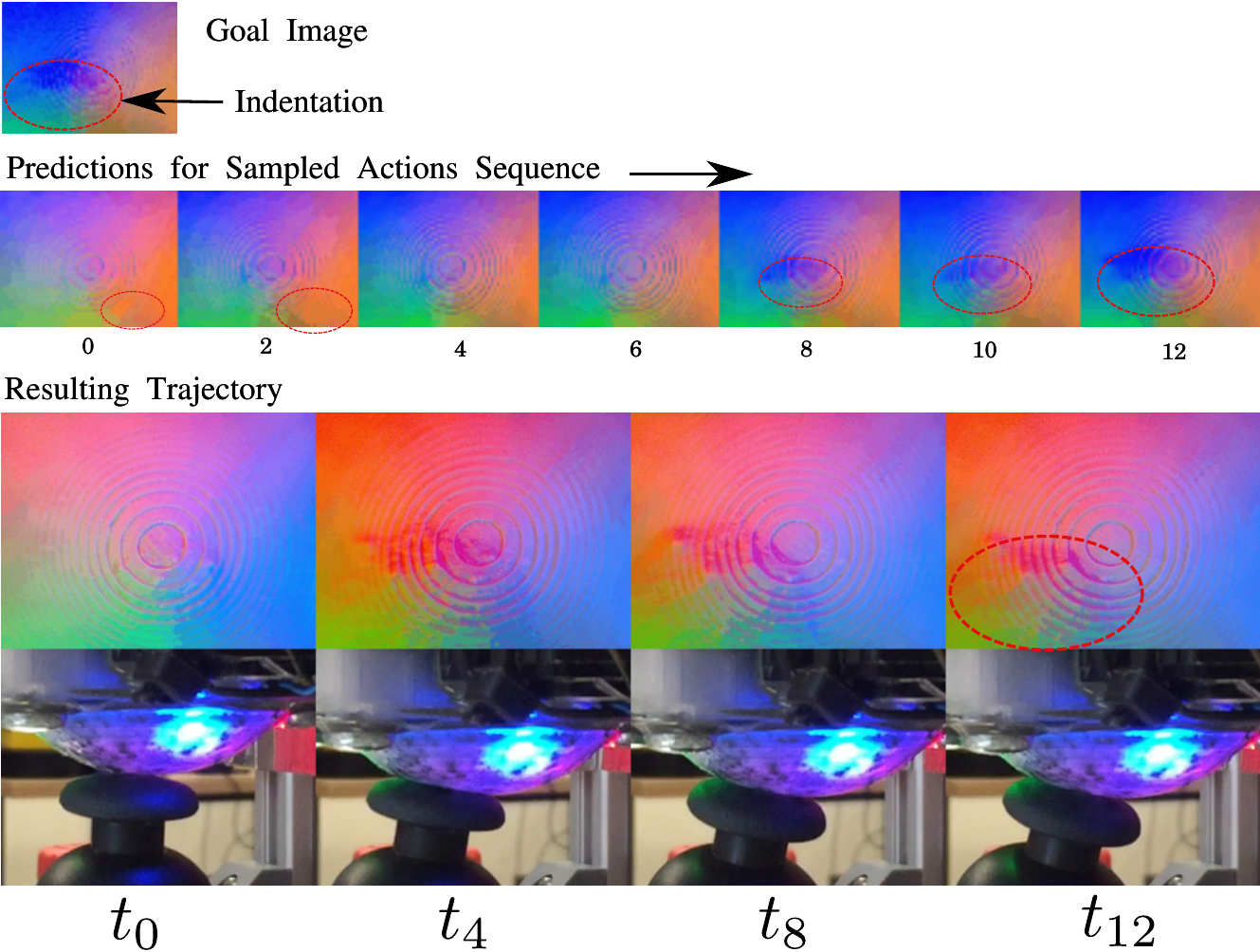} 
  \caption{Example of successful analog stick tactile positioning task. In the second row we show the predicted images (every 2nd time-step) for the optimal action sequence found at (real-world) timestep 1. For the 1st timestep (second row) the pressure center is in the bottom right of the image indicated by a red ellipse, it then lifts off for several timesteps and comes back in the last five steps. The last image of the predicted sequences closely resembles the desired indentation shown in the goal image.}
  \label{fig:analog_stick_result}
\end{figure}

\vskip 10mm

We now experimentally validate our deep tactile MPC approach on three real-world tactile manipulation tasks: moving a ball bearing, positioning an analog joystick, and rolling a die. For video results, see the project webpage\footnote{\url{https://sites.google.com/view/deeptactilempc}}.

\subsection{Evaluation Metrics}
Evaluating the performance of the tactile policy is challenging, since the target is provided directly in the space of tactile observations, and the ground truth pose of the object is unknown. However the advantage of using tactile images as goals is that this is highly general, since a user can easily specify any goal by manually positioning the sensor in the desired configuration.

For our evaluation, we use three different metrics that quantify different aspects of control performance: 1) mean squared error (MSE) of the difference between the goal-image $I_g$ and the tactile-image $I_t$ observed at the end of an episode, and  2) manually annotated distance in pixel-space between the pressure centroid of the object at the end of the trjectory and the location of the pressure centroid in the goal image. 3) For the die-rolling task we have a more intuitively meaningful success metric --- the fraction of trials in which the die could be rolled so that it has the desired face on top.

While the MSE metric can be automatically evaluated and exactly captures the objective we optimize for in tactile MPC, mean squared errors in image space do not necessarily reflect actual distances between object positions and poses. It is for this reason that we use the additional manually annotated distance measure.

\subsection{Tactile Control Baseline}
\label{subsec:baseline}

In order to provide a comparative baseline for our tactile MPC  method,  we designed an alternative method that uses hand-engineered image features to solve each of our three evaluation tasks. It accomplishes this by first detecting the pressure centre in the imprint of the ball, joy-stick or die and then moving in a straight line towards the target position. To estimate the coordinates of the pressure center in the current image and in the goal image $I_g$, it computes the weighted centroid of the pixels in the current image, where weights are squared pointwise differences between the current image and a blank ``background'' image from the sensor when it is not in contact with any object. This difference image outlines the contact region, and detecting its centroid roughly localizes the object of interest. 

Having detected an approximate position of the object in both the current and goal image, the baseline commands an action along the vector from the current estimated contact position to its estimated position in the goal image. The step length is tuned to achieve the maximum control performance in terms of the estimated distance between the current and desired object position. Note that this is a fairly strong baseline for the ball bearing task, since localizing the centroid provides a good indication for the location of the spherical ball. For the joystick and die-rolling task, this baseline fails frequently, yet it is hard to design a better baseline. In contrast, deep tactile MPC is more general and does not require manual tuning or knowledge about specific object mechanics. The deep dynamics model used in tactile MPC learns a variety of basic properties about the world purely from data -- such as that ball bearings remain in one piece and move opposite to the direction of movement of the sensor.

\begin{figure}[t]
  \centering
  \includegraphics[width=0.49\linewidth]{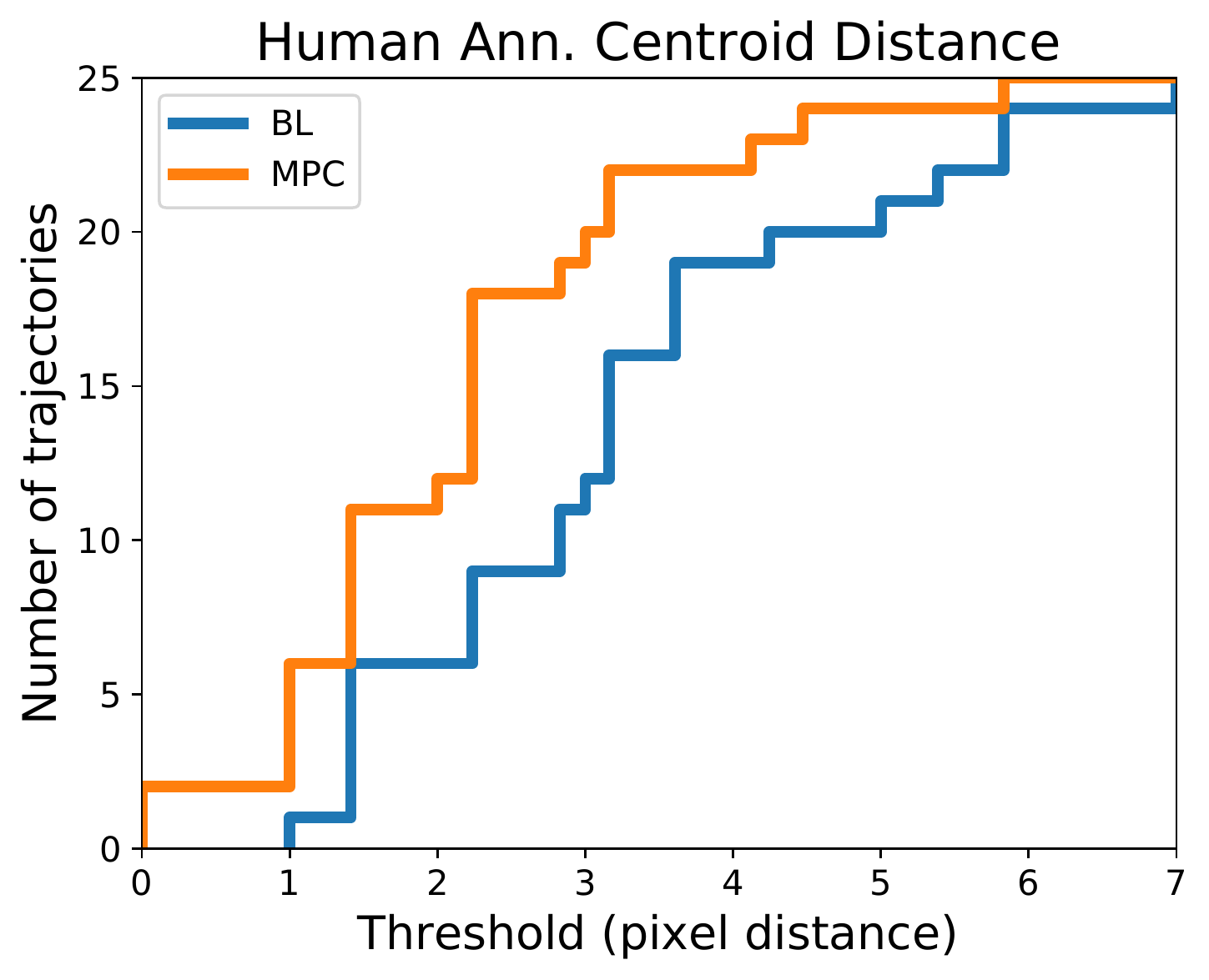}
  \hfill
  \includegraphics[width=0.49\linewidth]{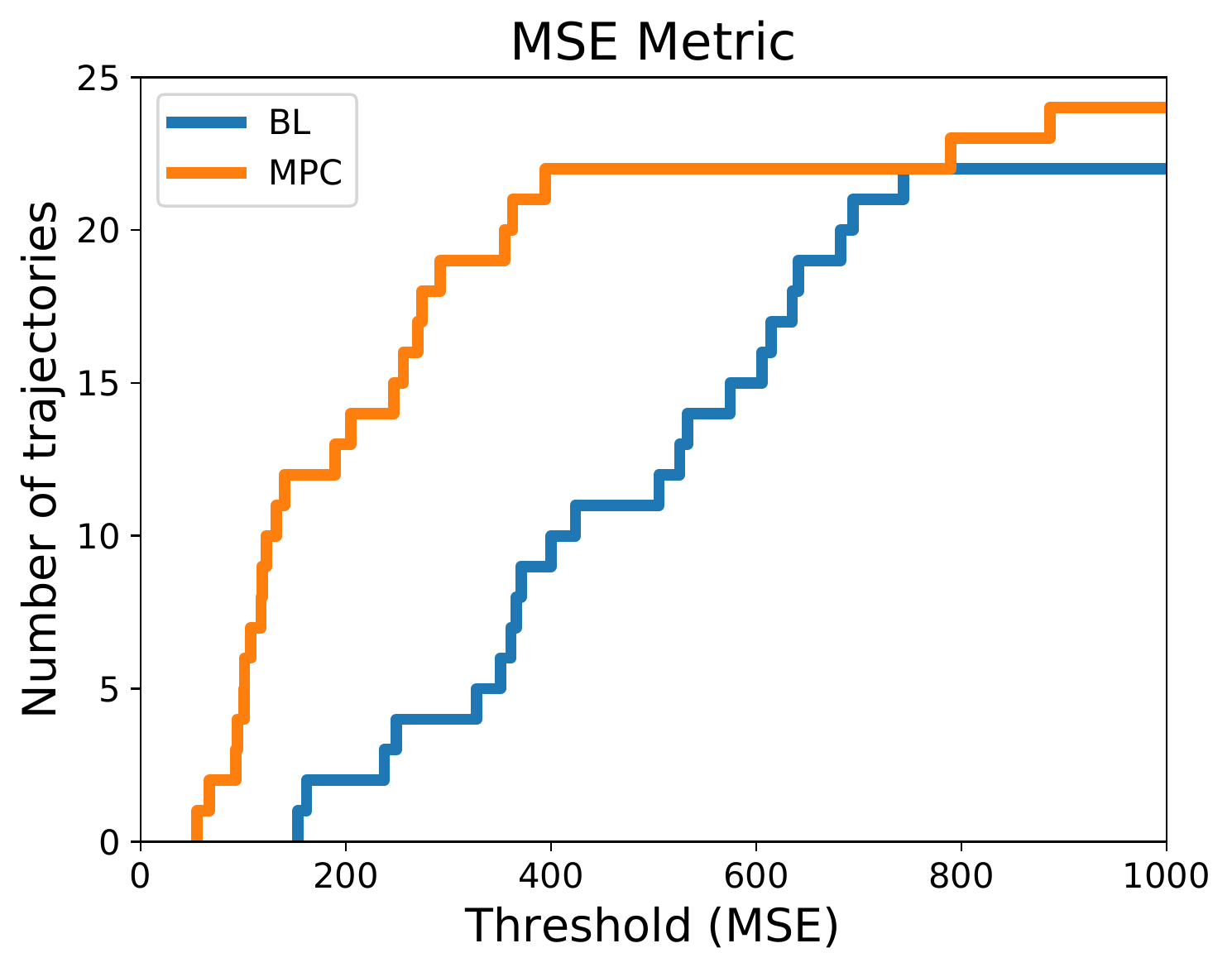}
  \caption{Quantitative analysis for ball task. (\textit{Left})~The y axis shows the number of trajectories out of 30 total for which the pixel distance between the final and the goal position of the pressure centroid, as annotated by a human labeler, is lower than the threshold given by the x-axis. (\textit{Right})~Number of trajectories with MSE distance to goal-image lower than threshold. A vertical slice through this can be used to determine what fraction of trajectories reach the goal within a certain distance. In all cases, our deep tactile MPC method (in orange) outperforms the hand-designed baseline approach (in blue) significantly.}
  \label{fig:results1}
\end{figure}

\subsection{Manipulating a Ball, an Analog Stick, and a 20-sided Die}

\begin{figure}[t]
  \centering
  \includegraphics[width=\linewidth]{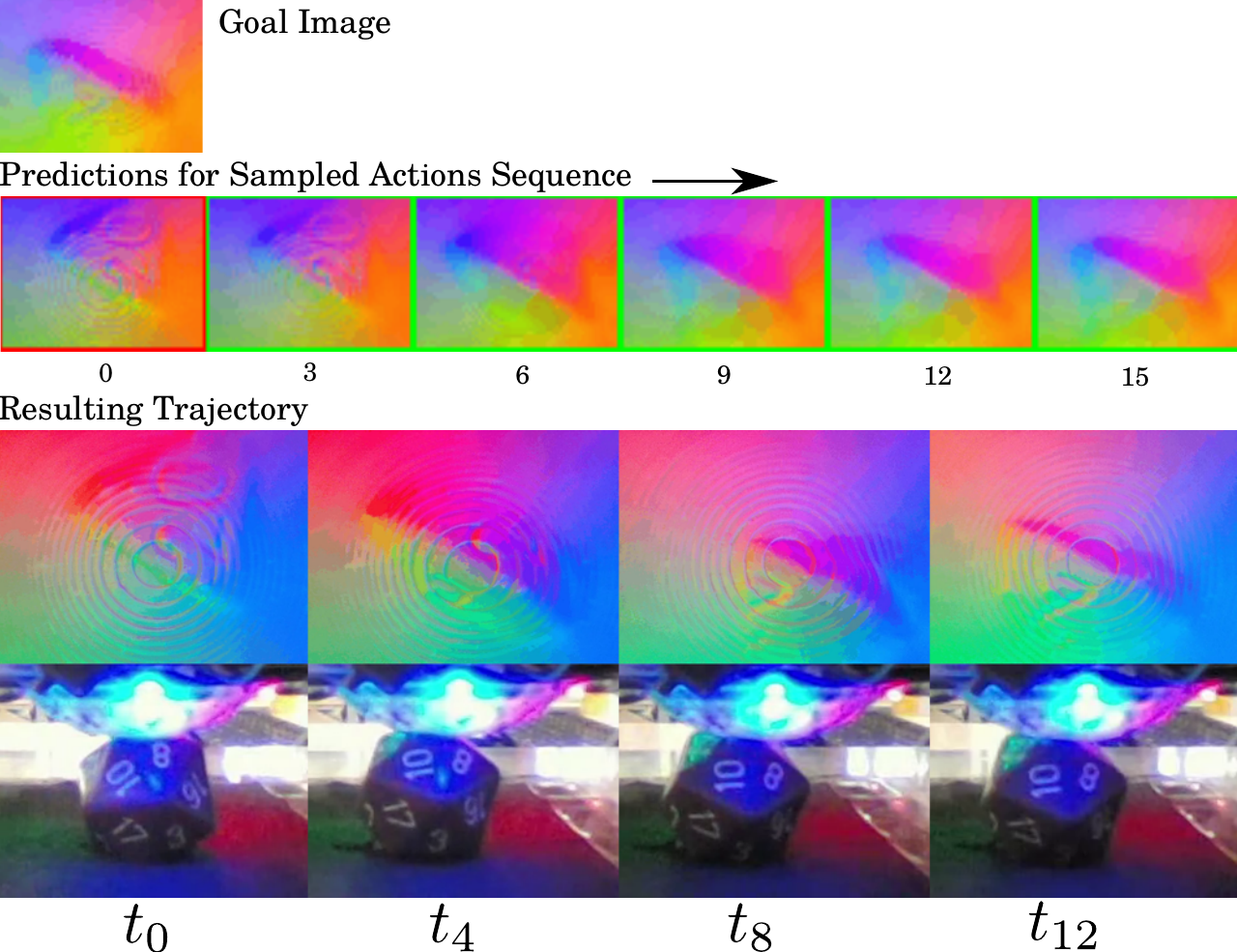}
  \caption{Example of successful execution of die rolling task. Starting from face 20 the goal is to reach face 8. The second row shows the video-predictions (at every 3rd time-step) for the best action sequence found at the first real-world time-step. The red margins indicate real context frames, green margins indicate predicted frames.}
  \label{fig:die_results}
\end{figure}

We find that our method enables a robot to achieve all three manipulation tasks through only touch sensing, without other sensory feedback. For example, the robot is able to maneuver a ball, manipulate a die to new faces, control a joystick, all entirely by feel. For qualitative examples of these results, 
see Figures \ref{fig:planning}, \ref{fig:analog_stick_result} and \ref{fig:die_results}, as well as the supplementary video\footnote{\url{https://sites.google.com/view/deeptactilempc}}.

\begin{table}[b]
\centering

\begin{tabular}{ | l | c | c | c | c }
    \hline
    & \multicolumn{2}{c|}{Median L2 dist [mm]} & Success Rate\\
    \cline{2-4}
     &  Ball Rolling & Analog Stick & Die \\ \hline
   Tactile MPC  &  \textbf{2.10} &  \textbf{5.31} & \textbf{86.6\% (26/30)} \\ \hline
   Centroid Baseline  &  2.97  & 8.86 & 46.6\% (14/30)   \\ \hline
\end{tabular}
\caption{Benchmark results for the ball-rolling, analog-stick and die-rolling experiments. The median L2 distances are between the hand-annotated pressure centroid of the final and goal-image. For the die experiment we measure the fraction of examples where the desired face lands on top. Benchmarks are performed with 30 examples.}
\label{tab:results}
\end{table}

\begin{figure}[t]
  \centering
  \includegraphics[width=0.49\linewidth]{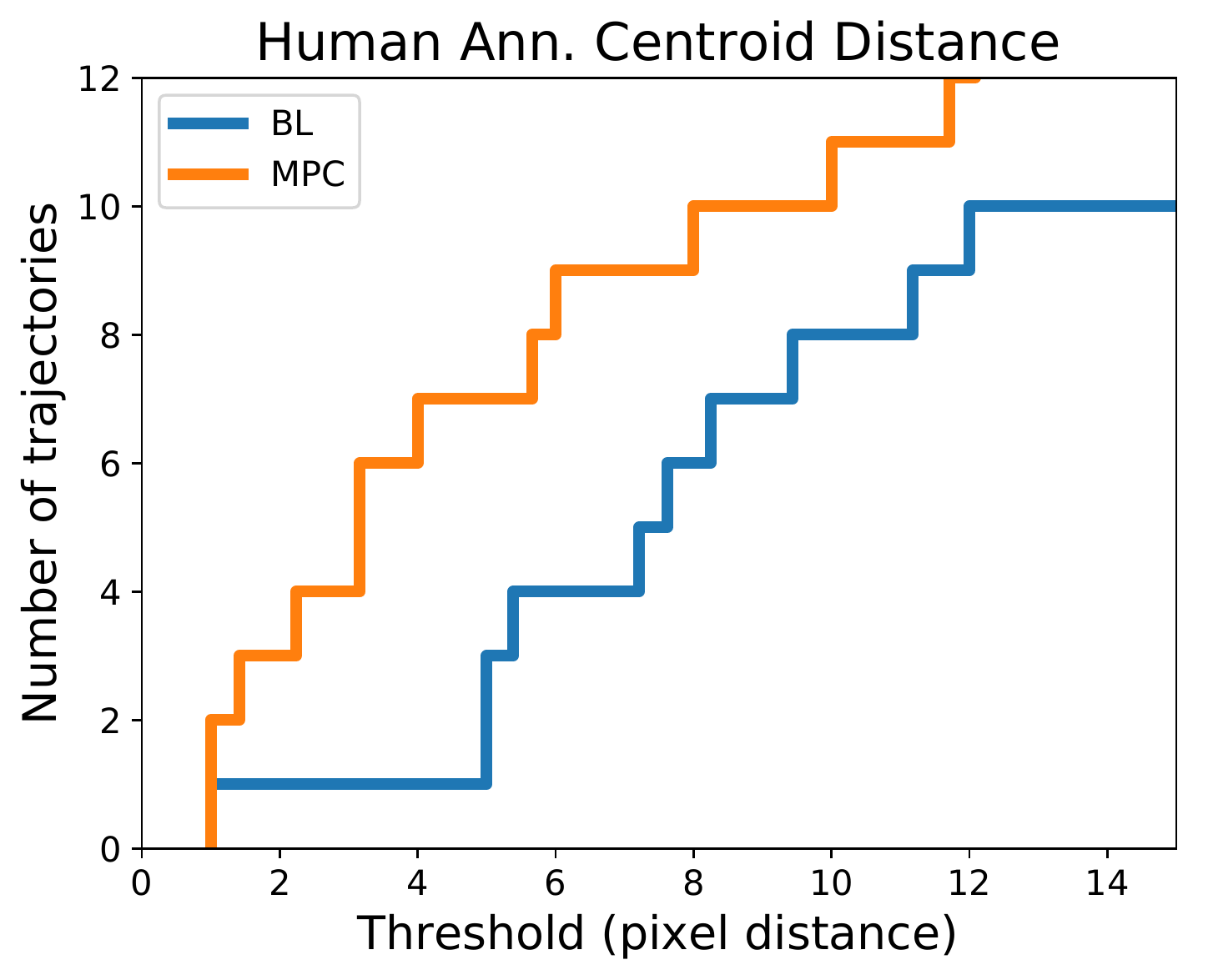}
  \hfill
  \includegraphics[width=0.49\linewidth]{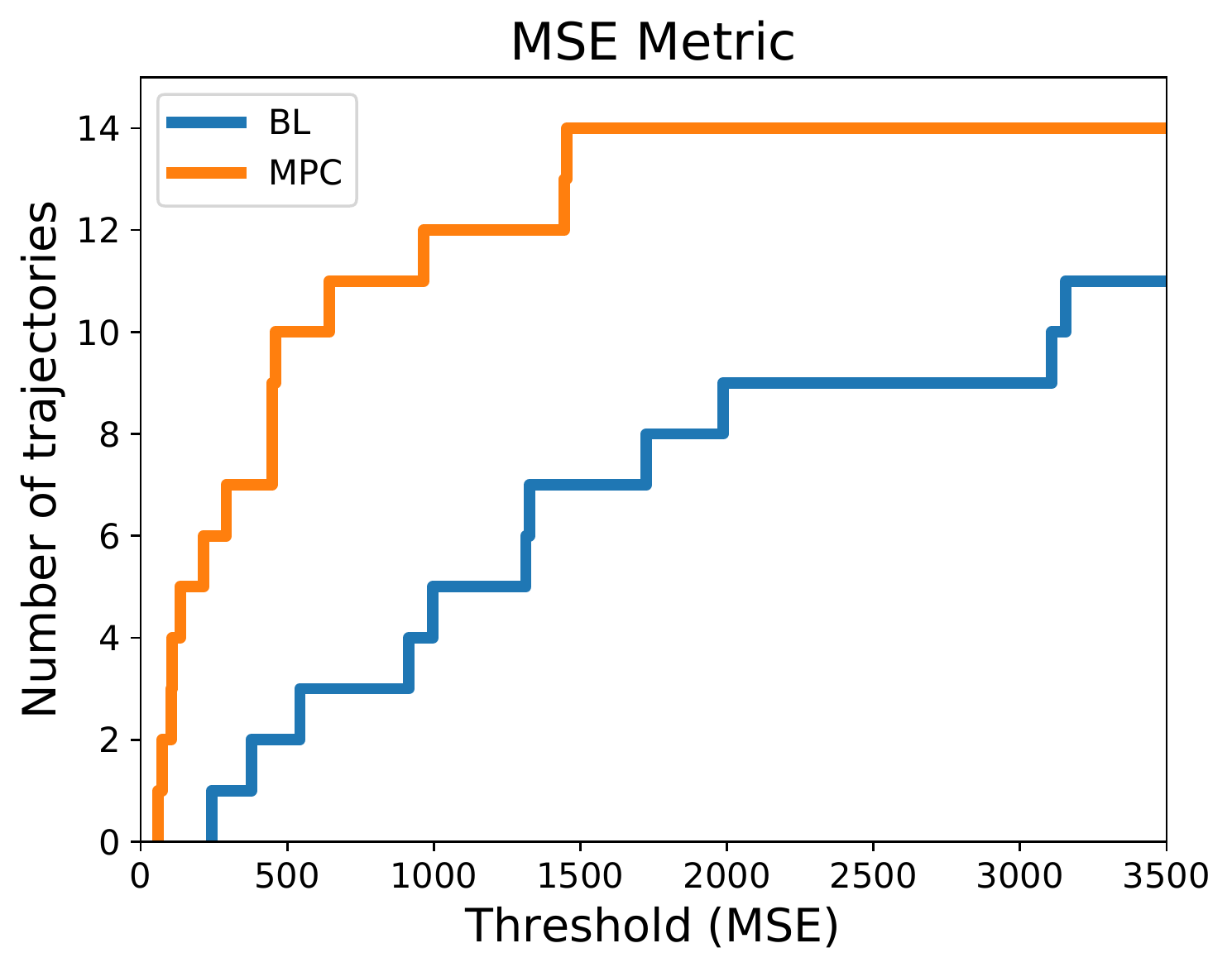}
    \caption{Quantitative analysis for analog-stick task. (\textit{Left})~Number of trajectories, out of 15 trials, for which euclidean distance between the final position of the pressure centroid and the goal position, as labeled by a human labeler. (\textit{Right})~Number of trajectories for which mean squared error (MSE) between the final image and the goal image is lower than threshold.}
  \label{fig:analog_stick_results}
\end{figure}

For the ball repositioning task, on the left of \autoref{fig:results1} we show a plot that illustrates the fraction out of 30 test trajectories that obtained distances between the pressure centroid at the final time-step and goal image which are lower than a certain threshold. The positions were hand-labeled, the distances are measured in terms of distances in the 64x48 tactile image. The right sight of \autoref{fig:results1} shows the same graph for the mean-squared error between the final image and the goal-image. Note that for both metrics our method (in orange) consistently dominates the baseline approach (in blue) by a substantial margin. 

The results for the analog stick repositioning task are shown in \fig{fig:analog_stick_results}, using the same metrics. Again, we see that our method (in orange) substantially outperforms the baseline.

As shown in \tab{tab:results}, our deep tactile MPC method achieves a significantly lower median distance than the baseline in both the ball-rolling and analog-stick task. In the die rolling experiments the difference between tactile MPC and the baseline is even larger. We conjecture that this is because of the fact that the dynamics in this die rolling task are too complex to be handled well by a simple hand-tuned controller, since it involves complex slipping, sliding and rolling motions on multiple surfaces. \autoref{fig:die_results} shows a qualitative example (more in supplementary video).

Based on these results we conclude that using sampling-based planning in a combination with a deep dynamics model is powerful method for solving a range of challenging manipulation tasks solely based on tactile information. We expect that the gap between hand-design methods and learning-based method will be even greater for more complex robotic manipulation scenarios such as multi-fingered manipulation.

\section{Discussion and Future Work}
\label{sec:conclusion}

Precise in-hand manipulation in humans heavily relies on tactile sensing. Integrating effective touch sensing into robotic control can enable substantially more dexterous robotic manipulation, even under visual occlusion and for small objects that are otherwise difficult to perceive. In this paper, we presented a touch-based control method based on learning forward predictive models for high-bandwidth GelSight touch sensors. Our method can enable a robotic finger to reposition objects and reach user-specified goals.

While our results indicate that deep convolutional recurrent models can effectively model future touch readings conditioned on a robot's actions, our method still has a number of limitations. First, we explore short-horizon control, where the goal can be reached using only tens of time steps. While this is effective for simple servoing tasks, it becomes limiting when tasks require rearranging multiple objects or repeatedly executing more complex finger gaits. However, as video prediction models improve, we would expect our method to improve also, and to be able to accommodate more complex tasks.

Another limitation of our work is that, with a single finger, the range of manipulation behaviors that can be executed is limited to simple rearrangement. A more dexterous arm or a multi-fingered hand could perform more complex manipulation tasks. An exciting direction for future work would be to extend our results with multiple fingers equipped with touch sensors, with a joint predictive model that can predict the dynamics of object interaction with all of the fingers at once. Such setting could perform complex in-hand object repositioning, assembly, and other manipulation skills.

\section*{Acknowledgements} 
We would like to thank Chris Myers from the CITRIS Invention Lab at UC Berkeley for help with building the custom GelSight sensor and the 3 axis test-rig.

\bibliographystyle{IEEEtran}
\bibliography{paper-tactile-servoing}  %

\clearpage
\section{Appendix}

\subsection{Autonomous Data Collection}
\label{sec:app_data_coll}
When performing autonomous data collection we either need to reset the environment to a well-defined set of starting conditions after each trajectory or the set of reachable states in the environment needs to be confined. In the case of the ball-rolling task we use slightly curved surface so that upon completion of a trajectory the ball automatically rolls back to a location close to the middle of the arena. For the analog-stick task a reset mechanism was provided by the springs embedded into the analog stick. In the die rolling task we used a thread fastened to the die wound onto a motor that resets the die to an approximately fixed starting pose at the beginning of each trial. For each trial, the sensor makes contact with the surface of the die and moves it in an arbitrary direction resulting in different die faces. At the end of each trial, the die is reset as described above.

To that end, we collected 7400 trajectories for the ball, around 3000 trajectories for the analog stick, and 4500 trajectories for the die experiment. Both during data collection and planning the actions are parameterized as changes in finger position of $\pm$ \SI {6.0}{\milli\meter} in the $x$, $y$, and $z$ directions. Data is collected at \SI{1.5}{\hertz}. At test time, tactile MPC runs at around 1 Hz. Both during data collection and planning we repeat actions for 3 time-steps, but we record images at every time-step, providing advantages for planning as explained in the paragraph \textit{Implementation details}. We found that having a higher frequency for images than actions helps the model making more accurate predictions in environments with discontinuous dynamics.

\subsection{Deep Recurrent Visual Dynamics Model}
\begin{figure}[t]
  \centering
  \includegraphics[width=0.98\linewidth]{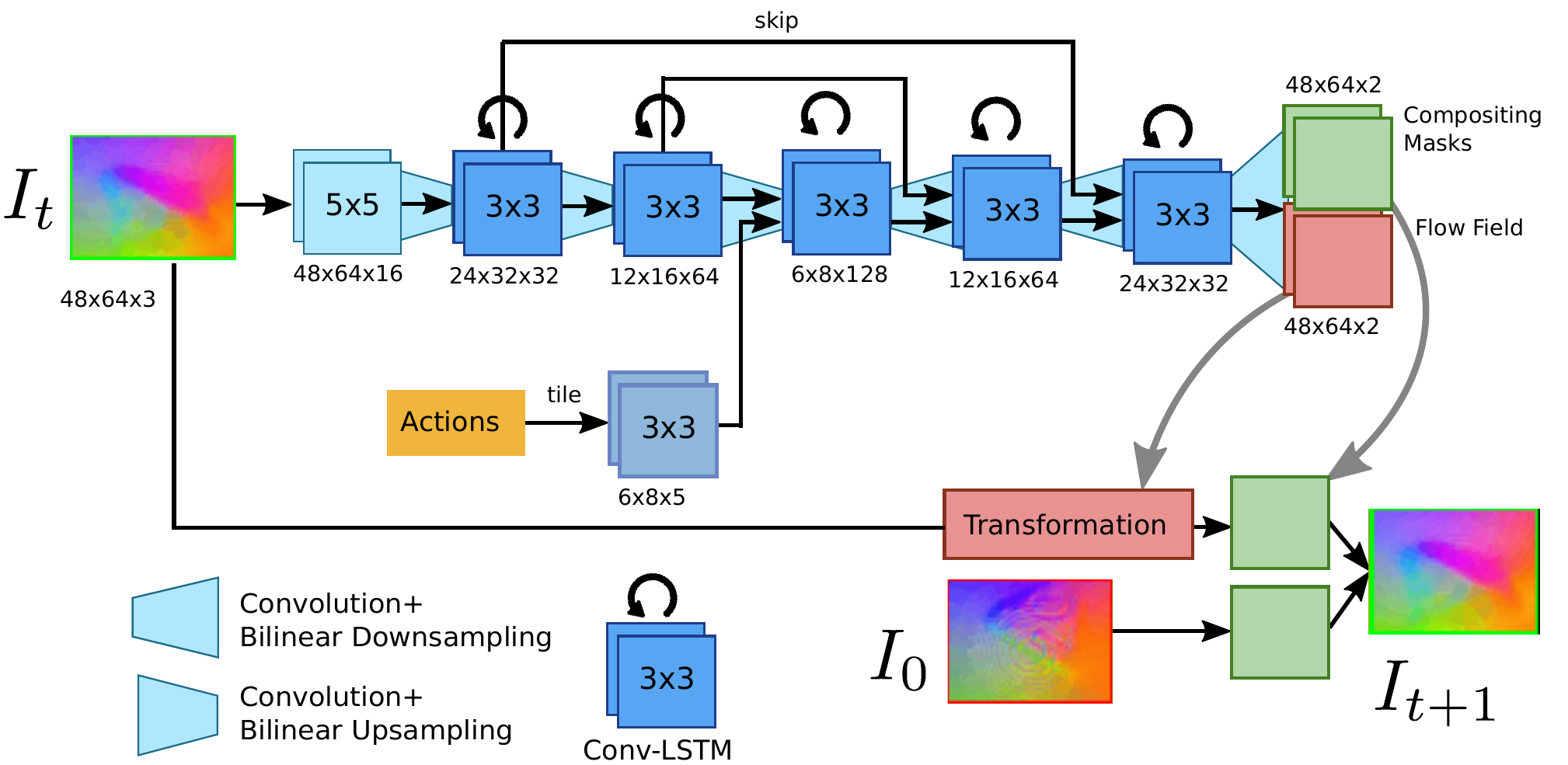}
  \caption{Video Prediction Architecture. }
  \label{fig:vidpred}
 \end{figure}
\label{sec:app_model}
The video prediction model is implemented as a deep recurrent neural network. Future images are generated by applying transformations to previous images. A schematic is shown in \autoref{fig:vidpred}. More details on the video-prediction architecture can be found in \cite{Ebert2017} and \cite{savp}. Note that depending on the experiment during the first 3 time-steps of unrolling the RNN prediction model we feed the most recent ground truth observations, we call these images \emph{context frames}.

\end{document}